\documentclass[letterpaper, 10 pt, conference]{ieeeconf}  

\usepackage{cite}
\usepackage{amsmath,amssymb,amsfonts}
\usepackage{algorithmic}
\usepackage{graphicx}
\usepackage{textcomp}
\usepackage{xcolor}
\usepackage{caption}
\usepackage{subcaption}

\makeatletter
\def\maketag@@@#1{\hbox{\m@th\normalfont\normalsize#1}}
\makeatother

\usepackage[linesnumbered,ruled,vlined]{algorithm2e}

\newtheorem{theorem}{Theorem}

\newtheorem{corollary}{Corollary}

\newcommand{\DDABSP}[1]{{\color{magenta} #1}}

\newcommand{\prob}[1]{\ensuremath{\mathbb{P}({#1})}}

\IEEEoverridecommandlockouts                              

\title{\LARGE \bf
D2A-BSP: Distilled Data Association Belief Space Planning with Performance Guarantees Under Budget Constraints
}

\author{Moshe Shienman and Vadim Indelman
\thanks{Moshe Shienman is with the Technion Autonomous Systems Program (TASP), Technion - Israel Institute of Technology, Haifa 32000,
	Israel, {\tt smoshe@campus.technion.ac.il}. Vadim Indelman is with the Department of Aerospace Engineering, Technion - Israel Institute of Technology, Haifa 32000, Israel. {\tt vadim.indelman@technion.ac.il}. This work was  partially supported by US NSF/US-Israel BSF.}%
}

\begin{document}

\maketitle
\thispagestyle{empty}
\pagestyle{empty}

\begin{abstract}
Unresolved data association in ambiguous and perceptually aliased environments leads to multi-modal hypotheses on both the robot's and the environment state. To avoid catastrophic results, when operating in such ambiguous environments, it is crucial to reason about data association within Belief Space Planning (BSP). However, explicitly considering all possible data associations, the number of hypotheses grows exponentially with the planning horizon and determining the optimal action sequence quickly becomes intractable.
Moreover, with hard budget constraints where some non-negligible hypotheses 
must be pruned, achieving performance guarantees is crucial. 
In this work we present a computationally efficient novel approach that utilizes only a distilled subset of hypotheses to solve BSP problems while reasoning about data association. Furthermore, to provide performance guarantees, we derive error bounds with respect to the optimal solution. 
We then demonstrate our approach in an extremely aliased environment, where we manage to significantly reduce computation time without compromising on  the quality of the solution.     
\end{abstract}

\section{INTRODUCTION}
Decision making under uncertainty is at the core operation of intelligent autonomous agents and robots. Autonomous navigation, robotic surgeries and automated warehousing are only a few examples where agents must autonomously plan and execute their actions while reasoning about uncertainty. Such uncertainty might be due to noisy or limited observations; imprecise delivery of actions; or dynamic environments, where unpredictable events might take place. As the true state of the agent and the environment is unknown, it is represented by a probability density function (belief) over the corresponding random variables. To autonomously determine which actions to take, planning and decision making are performed over that distribution of possible states (the belief space). This problem is also known as Belief Space Planning (BSP) and is an instantiation of a Partially Observable Markov Decision Problem (POMDP) \cite{Kaelbling98ai}.

In ambiguous and perceptually aliased environments, BSP is even more challenging. In such scenarios, data association cannot be considered to be given and perfect, i.e. considering that the agent properly perceives the environment by its sensors, as it can lead to incorrect posterior beliefs and catastrophic results. As such, BSP should reason about data association while also considering other sources of uncertainty. However, reasoning about data association, the number of hypotheses grows exponentially with the planning horizon and determining the optimal action quickly becomes intractable. 
Moreover, when considering 
real time operation, using inexpensive hardware, hard computational budget constraints are often required, e.g. bounding the number of hypotheses supported in planning. As such, some non-negligible hypotheses might be pruned and achieving some performance guarantees becomes vital.

Given a set of candidate actions, the main goal of BSP is to retrieve the optimal action with respect to a user defined objective function. Specifically, in the case of a multi-modal belief (corresponding to different hypotheses), a traditional solution requires evaluating the objective function with respect to each hypothesis. Instead, we suggest to solve a simplified problem using only a distilled subset of hypotheses where the loss in solution quality can be bounded to provide performance guarantees. 

Our contributions in this paper are as follows: (a) We introduce a novel approach, \DDABSP{D2A-BSP}, that utilizes a distilled subset of hypotheses in planning to reduce  computational complexity; (b) We develop the connection between our approach and the true analytical solution, owing to every possible data association, for the myopic case; 
(c) We derive bounds over the true analytical solution, which can be incrementally adapted, and prove their convergence. Moreover, in a budget free scenario, these bounds are used to speed-up calculations while preserving  the same action selection as when considering all hypotheses; (d) Crucially, we address also the challenging setting of data association aware BSP with hard budget constraints, and show, for the first time, these bounds provide performance guarantees;
(e) We demonstrate the merits of our approach in a highly ambiguous scenario containing identical landmarks.

This paper is accompanied with supplementary
material \cite{Shienman22icra_Supplementary} which provides further details and results.

\section{RELATED WORK}
In an attempt to ensure reliable and efficient operation in ambiguous environments, different approaches were recently proposed. These approaches, often referred to as robust perception, usually maintain probabilistic data association and hypothesis tracking, given accessible data. 

With respect to the state inference problem, and specifically in the context of Simultaneous  Localization And Mapping (SLAM), the inference mechanism should be resilient to false data association overlooked by front-end algorithms, e.g. laser scans and image matching. A common approach to handle this problem is utilizing efficient graph representations.  
In \cite{Carlone14iros} the authors propose using a pose graph model, where graph optimization is efficient in finding the maximal subset of measurements that is internally coherent to discard false data associations. 
The authors of \cite{Indelman14icra} and \cite{Indelman16csm} utilized the same graphical model with an expectation-maximization (EM) approach to efficiently infer robot initial relative poses and solve a multi robot data association problem.
In \cite{Sunderhauf12icra} the authors modified parts of the topological structure of the graph during optimization to discard false positive loop closures.
In \cite{Olson13ijrr} the authors utilized factor graph \cite{Kschischang01it} representations to perform inference on networks of mixtures while in \cite{Hsiao19icra} the authors extended the Bayes tree \cite{Kaess12ijrr} algorithm to explicitly incorporate multi-modal measurements within the graph and generate multi-hypothesis outputs. Yet, all of these works were developed for the passive case only,  i.e.~no planning is involved.

Only recently, ambiguous data association was considered in different BSP approaches for active disambiguation. 
In \cite{Agarwal16wafr} the authors considered data association hypotheses within the prior belief, modeling it as a mixture of Gaussians, and assumed that at least one action can lead to complete disambiguation. However, their work does not reason about ambiguous data association within future beliefs (owing to future observations). The authors of \cite{Pathak18ijrr} incorporated, for the first time, reasoning about future data association hypotheses within a belief space planning framework, terming the corresponding approach DA-BSP. Another related work in this context is \cite{Hsiao20iros}, that also reasons about  ambiguous data association in future beliefs while utilizing the graphical model presented in \cite{Hsiao19icra}.  
To handle the exponential growth in the number of hypotheses, these approaches suggested to use different heuristics, e.g. pruning and merging. However, none of them developed any analytical bounds on the loss in quality of the solution, with respect to the original problem, and cannot provide performance guarantees.  

While finding the optimal solution of a POMDP was proven to be computationally intractable \cite{Papadimitriou87math}, several approaches were developed over the years to reduce  computational complexity and allow online operation while planning under uncertainty. 
Some methods rely on approximated solutions via direct trajectory optimization, e.g. \cite{Indelman15ijrr} and \cite{VanDenBerg12ijrr}, while others approximate the state or the objective function to reduce the planning complexity, e.g. \cite{Bopardikar16ijrr}.
Belief sparsification in planning was first introduced in \cite{Indelman16ral} to limit the state size and allow long-term operation. The author utilized a diagonal covariance approximation, in a myopic setting with one-row unary Jacobians, to maintain a similar action selection while significantly reducing the complexity of the objective calculation.
The authors of \cite{Elimelech21ijrr} presented a sparsification approach to handle BSP problems. They suggested to identify uninvolved variables and sparsify the posterior information matrix for each candidate action to reduce computation time. Other recent approaches suggest utilizing structural properties of different graphical models in decision making under uncertainty, e.g. in \cite{Kitanov18icra},\cite{Kitanov19arxiv} and \cite{Shienman21ral} different topological signatures were used to approximate the solution to BSP problems. 
Yet, in all of these approaches data association is assumed the be known and perfect.

To the best of our knowledge, in-spite of aforementioned research efforts, simplifying the BSP problem while reasoning about data association and maintaining performance guarantees is a novel concept.

\section{NOTATIONS AND PROBLEM FORMULATION} \label{notations section}
Consider an autonomous agent operating in a known environment, where different objects or scenes can possibly be perceptually similar or identical. The agent aims to decide its future actions, while reasoning about ambiguous data association, based on information accumulated thus far and a user defined objective function.  

\subsection{Belief Propagation}
Let $x_k$ denote the agent's state at time instant $k$.  
For simplicity, in this work we assume the environment, represented by landmarks, to be given. We note that extending our approach to a full SLAM scenario is straightforward, using similar notations as in \cite{Tchuiev19iros}. 

We denote the data association realization vector at time $k$ as $\beta_k$. Given $n_k$ observations at time $k$, $\beta_k \in \mathbb{N}^{n_k}$. Elements in $\beta_k$ are associated according to the given observation model and each element, i.e. landmark, is given a unique label. A specific data association hypothesis at time $k$ is thus given by a specific set $j$ of associations up to and including time $k$ and is denoted as $\beta_{1:k}^j$.

Let $Z_k \triangleq \lbrace z_{k,1},...,z_{k,n_k} \rbrace$ denote the set of all $n_k$ measurements at time $k$ and let $u_k$ denote the agent's action at time $k$. $Z_{1:k}$ and $u_{0:k-1}$ denote all observations and actions up to time $k$,  respectively. The motion and observation models are given by
\begin{equation} \label{eq:motion and observation models}
	x_{k+1} = f \left( x_k,u_k,w_k \right) \quad , \quad z_{k}=h \left( x_k, l, v_k \right),
\end{equation} 
where $l$ is a landmark pose, and $w_k$ and $v_k$ are noise terms, sampled from known motion and measurement distributions,  respectively.

The posterior probability density function (pdf) over the state $x_k$, denoted as the \textit{belief}, is given by 
\begin{equation} \label{eq:joint belief}
	b \left[ x_k \right] \triangleq \mathbb{P} \left( x_k |z_{0:k}, u_{0:k-1}\right) = \mathbb{P} \left( x_k|H_k\right) ,
\end{equation}
where $H_k \triangleq \lbrace Z_{1:k}, u_{0:k-1} \rbrace$ represents history at time $k$.
We define $H_{k+1}^{-} \triangleq H_k \cup \lbrace u_k \rbrace$ and $b_{k+1}^{-} \triangleq \mathbb{P} \left( x_{k+1} | H_{k+1}^{-} \right)$ for notational convenience. The belief at time $k$ is denoted from hereon as $b_k$.

As data association is not given, and different observations might be attributed to different but similar-in-appearance landmarks, the belief at time $k$ includes different hypotheses.
In particular, marginalizing over $M_k \in \mathbb{N}$ hypotheses and using the chain rule, we rewrite the belief at time $k$ as a linear combination 
\begin{equation} \label{belief mixture}
	b_k  = \sum_{j=1}^{M_k} 
	\underbrace{\mathbb{P} \left( x_k|\beta_{1:k}^j, H_k\right)}_{b^j_k} \underbrace{\mathbb{P} \left( \beta_{1:k}^j | H_k\right)}_{w_k^j},
\end{equation}
where $b^j_k$ is a conditional belief, with some general distribution, that corresponds to the $j$th hypothesis, and $w_k^j$ is the associated weight.

Updating the belief, after performing control $u_{k+1}$ and taking an observation $Z_{k+1}$, also requires reasoning about data association. Given $M_k$ hypotheses from time $k$, marginalizing over all landmarks at time $k+1$ and using the chain rule we explicitly write it as
\smaller
\begin{equation} \label{propagated belief with weights}
	b_{k+1}  = \sum_{i=1}^{|L|} \sum_{j=1}^{M_k} \underbrace{\mathbb{P} \left(x_{k+1} |   H_{k+1}, \beta_{k+1}^i, \beta_{1:k}^j \right)}_{b^{i,j}_{k+1}} \underbrace{\mathbb{P} \left( \beta_{k+1}^i, \beta_{1:k}^j | H_{k+1}\right)}_{w^{i,j}_{k+1}},
\end{equation}
\normalsize
where $|L|$ represents the number of different data association realizations considered at time $k+1$.
The first term $b^{i,j}_{k+1}$ represents a conditional belief at time $k+1$ which originated from the $j$th hypothesis at time $k$ and a specific data association realization $\beta_{k+1}^i$. The second term $w^{i,j}_{k+1}$ is the associated belief component weight.

\begin{corollary} \label{corollary 1}
\textit{Each posterior belief component weight $w^{i,j}_{k+1}$ can be written as}
\begin{equation}	
w^{i,j}_{k+1} = \eta_{k+1}^{-1} \tilde{\zeta}_{k+1}^{i,j} w_k^j,
\label{eq:updated_weights}
\end{equation}
\textit{where $w_k^j$ is the weight of the $j$th component from time $k$; $\eta_{k+1}$ is a normalization term; and $\tilde{\zeta}_{k+1}^{i,j}$ 
is the probability for the $i$th data association at time $k+1$ given the $j$th hypothesis from time $k$},
	\begin{equation} \label{eq:zeta i,j}	
	\tilde{\zeta}_{k+1}^{i,j} \triangleq \mathbb{E}_{x_{k+1}}[\mathbb{P}\left( Z_{k+1}|\beta_{k+1}^i, x_{k+1} \right) \mathbb{P} \left( \beta_{k+1}^i|x_{k+1} \right) ], 
\end{equation}
\textit{where the expectation is with respect to  \small $\mathbb{P} \left(x_{k+1}|H_{k+1}^{-},\beta_{1:k}^j \right)$. \normalsize }
\end{corollary}
The term $\mathbb{P}\left( Z_{k+1}|\beta_{k+1}^i, x_{k+1} \right)$ in \eqref{eq:zeta i,j} is the joint measurement likelihood for all observations obtained at time $k+1$ given the $i$th data association and state $x_{k+1}$. It can be explicitly written as
\begin{equation} \label{eq:joint measurement likelihood}
	\mathbb{P}\left( Z_{k+1}|\beta_{k+1}^i, x_{k+1} \right) = \!\! \prod_{r=1}^{n_{k+1}} \! \mathbb{P}\! \left( z_{k+1,r}|l_{\beta_{k+1}^i \left(r\right)}, x_{k+1} \right),
\end{equation}
where $l_{\beta_{k+1}^i \left(r\right)}$ denotes the landmark pose, corresponding to the $r$th measurement in the given data association realization vector $\beta_{k+1}^i$.

To allow fluid reading, proofs for all corollaries and theorems are given in the appendix.

\subsection{Belief Space Planning} \label{fomulation bsp}

Let $J$ denote a user defined objective function given by
\begin{equation} \label{eq:general objective}
	J \left(b_k, u_{k:k+N-1} \right) \! = \! \underset{}{\mathbb{E}}\!\left[ \sum_{n=1}^{N} \! c_{k+n} \left( b_{k+n}, u_{k+n-1} \right) \right],
\end{equation}
where $c_{k+n}$ represents the cost function associated with the $n$th look-ahead step and where the expectation is taken with respect to future observations $Z_{k+1:k+N}$.

Given a set of candidate action sequences $\mathcal{U}$  and a belief $b_k$, the goal of BSP is to find the optimal action sequence given by
\begin{equation} \label{optimization objective}
	u_{k:k+N-1}^* = \underset{\mathcal{U}}{\text{argmin}} \, J \left( b_k, u_{k:k+N-1} \right).
\end{equation}
Evaluating \eqref{optimization objective} at each planning session for every candidate action sequence is known to be computationally intractable even without reasoning about data association. Using \eqref{propagated belief with weights} it is not hard to see that  
explicitly reasoning about data association in planning adds an additional complexity as the number of belief components $|M_k| \left|L \right|^n$ grows exponentially with the planning horizon.

To relax the computational complexity, one could consider solving a computationally easier, simplified problem, with respect to the same set of candidate actions. If the solution can be mathematically related to the solution of the original problem, one can provide performance guarantees. This can be achieved by simplifying and bounding any of the objective function terms. 

\section{APPROACH}
We propose a method that reduces the computational complexity of BSP problems in which ambiguous data association is explicitly considered while providing performance guarantees on the quality of the solution.

As a first step towards applying our method for the general BSP problem \eqref{optimization objective}, in this work we consider a myopic setting, i.e. one look-ahead step, which by itself can be computationally challenging in highly ambiguous scenarios. Writing the expectation operator in \eqref{optimization objective} explicitly, the objective function for the myopic setting is defined as
\begin{equation} \label{eq:myopic objective}
	J \left(b_k, u_k \right) = \int\displaylimits_{Z_{k+1}} \eta_{k+1}
	c \left( b_{k+1} \right) dZ_{k+1},
\end{equation}
where $\eta_{k+1} \triangleq \prob{Z_{k+1} | H_{k+1}^{-}}$ is the joint measurement likelihood, denoted from hereon simply as $\eta$.  
In this work we interchangeably refer to $\eta$ as the normalization term and the measurement likelihood.

In our approach we suggest using only a distilled subset of belief components $M_k^s \subseteq M_k$ from time $k$. We avoid calculating the posterior belief at time $k+1$ for components we do not consider in $M_k^s$. As such, the number of belief components at time $k+1$ reduces from $|M_k| \left| L \right|$ to $|M_k^s| \left| L \right|$
which also lowers the computational complexity of the considered cost function. When committing to a certain computational budget $Q$ over the number of posterior belief components, the distilled subset $M_k^s$ is subject to 
$|M_k^s| \left| L \right| \leq Q$. In this work, we only control the size of $M_k^s$. Crucially, we analytically bound the loss in solution quality for every considered action with respect to \eqref{eq:myopic objective}.  

We formally define a simplified belief at time $k$ as
\begin{equation} \label{eq: simplified belief mixture}
	b^s_k \triangleq  \sum_{j=1}^{M_k^s} w_k^{s,j} b^j_k \quad , \quad w^{s,j}_{k} \triangleq \frac{w^{j}_{k}}{w_k^{m,s}},
\end{equation}
where weights are re-normalized with $w_k^{m,s} \triangleq \sum_{m \in M^{s}_{k}} w^{m}_{k}$.

To provide performance guarantees, we wish to bound \eqref{eq:myopic objective}, for each candidate action $u_k$, using $b_k^s$
\begin{equation} \label{eq:Obj_bounds}
	\underline{J} \left( b_k, b_k^s, u_k \right) \leq
	J \left( b_k, u_k \right) \leq
	\bar{J} \left( b_k, b_k^s, u_k \right).	
\end{equation}
To efficiently evaluate these bounds, in this work we consider simplifying and analytically bounding both $\eta$ and the cost function terms in \eqref{eq:myopic objective}. Thus, we rewrite \eqref{eq:Obj_bounds} as
\scriptsize
\begin{equation} \label{eq: obj explicit bounds}
	\!\!\! \int\displaylimits_{Z_{k+1}} \!\!\!\!\!\mathcal{LB} \left[ \eta \right] \mathcal{LB} \left[ c \left( b_{k+1}\right) \right] dZ_{k+1} \!\! \leq \! 
	J \left( b_k, u_k \right) \! \leq \!\!\!\!\!
	\int\displaylimits_{Z_{k+1}} \!\!\!\!\!\mathcal{UB} \left[ \eta \right] \mathcal{UB} \left[ c \left( b_{k+1}\right) \right]dZ_{k+1}\!,
\end{equation} \normalsize
where $\mathcal{LB}$, $\mathcal{UB}$ denote lower and upper bounds, respectively. By definition, if $\mathcal{LB} \left[ \eta \right], \mathcal{UB} \left[ \eta \right]$ converge to $\eta$ and $\mathcal{LB} \left[ c \left( b_{k+1}\right) \right], \mathcal{UB} \left[ c \left( b_{k+1}\right) \right]$ converge to $c \left( b_{k+1}\right)$, the bounds in \eqref{eq: obj explicit bounds} converge to $	J \left( b_k, u_k \right)$.

\subsection{Bounding the cost function}
While the cost function in \eqref{eq:myopic objective} can generally include a number of different terms, e.g. distance to goal, energy spent and information measures of future beliefs, in this work we only consider an information theoretic term over data association hypotheses weights that can be used for autonomous active disambiguation of hypotheses. We believe that conceptually similar derivations can also support other terms, e.g. distance to goal, and leave that for future research.

Specifically, to disambiguate between hypotheses, we utilize the Shannon entropy, defined as $\mathcal{H} \triangleq - \sum\limits_{i=1}^{n} w^i log \left( w^i \right)$, where each $w^i$ corresponds to a belief component weight and $\sum\limits_{i=1}^{n} w^i = 1$. Using Corollary \ref{corollary 1}, we rewrite $\mathcal{H}$ as
\begin{equation} \label{entropy full}
	c\left( b_{k+1} \right) \triangleq \mathcal{H} = - \sum_{i}^{|L|} \sum_{j}^{M_{k}} \frac{  \tilde{\zeta}^{i,j}_{k+1}  w^{j}_{k}}{\eta} log \left(\frac{  \tilde{\zeta}^{i,j}_{k+1}  w^{j}_{k}}{\eta} \right).
\end{equation} 
To bound this cost function given a belief $b_{k+1}$  using the same cost function given a simplified belief $b_{k+1}^s$, we first rigorously derive the analytic connection between the two. \vspace{2pt}
\begin{theorem} \label{analytic cost theorem}
	\textit{Given a simplified belief $b_k^s$ at time $k$, for every action $u_k$ and considered future observation $Z_{k+1}$, the cost due to ambiguity 
		\eqref{entropy full} can be expressed by}
	\smaller
	\begin{multline} \label{eq:theorem 3}
		\mathcal{H} = \frac{ w_k^{m,s}}{\eta} \left[ \eta^s \left[ \mathcal{H}^s - log (\eta^s) \right]
		- \sum_{i}^{|L|} \sum_{j}^{M_{k}^s}  \tilde{\zeta}^{i,j}_{k+1} w^{s,j}_{k} log \left( \frac{w_k^{m,s}}{\eta} \right) \right] \\
		- \sum_{i}^{|L|} \sum_{j}^{\neg M^{s}_{k}} \frac{\tilde{\zeta}^{i,j}_{k+1}  w^{j}_{k}}{\eta}  log \left(\frac{\tilde{\zeta}^{i,j}_{k+1} w^{j}_{k}}{\eta} \right), 
	\end{multline}
	\normalsize
	\textit{where  \smaller $\neg M_k^s \triangleq M_k \setminus M_k^s$; $\mathcal{H}^s \triangleq c \left( b_{k+1}^s \right)$; and $\eta^s \triangleq \mathbb{P} \left( Z_{k+1} | b_{k}^{s}, u_k \right)$. \normalsize}
\end{theorem}

We now use Theorem \ref{analytic cost theorem} to derive bounds for $\mathcal{H}$ 
which are computationally more efficient to calculate as we only consider a subset of hypotheses. As can be seen in  \eqref{eq:eta explicit} and Section 4.1 in \cite{Pathak18ijrr}, evaluating $\eta$ requires evaluating all posterior components weights $w_{k+1}^{i,j}$. As our considered cost is a function of these weights, simplifying and bounding $\mathcal{H}$ has no computational merits without simplifying and bounding $\eta$   (denoted below by $\eta^s$, $\mathcal{LB}\left[{\eta}\right]$ and $\mathcal{UB}\left[{\eta}\right]$).  \vspace{2pt}
\begin{theorem} \label{cost bounds theorem}
	\textit{Given a simplified belief $b_k^s$ at time $k$, the cost due to ambiguity 
		term in \eqref{eq:myopic objective} is bounded by}
	\small
	\begin{multline} \label{eq:entropy lower bound}
		\mathcal{LB} \left[ c \left( b_{k+1}\right)\right] \triangleq \mathcal{LB} \left[ \mathcal{H} \right] =   \frac{\eta^s w_k^{m,s}}{\mathcal{UB}\left[{\eta}\right]} \left[\mathcal{H}^s - log (\eta^s) \right] \\
		- \frac{w_k^{m,s}}{\mathcal{UB}\left[{\eta}\right]} \sum_{i}^{|L|} \sum_{j}^{M_{k}^s}  \tilde{\zeta}^{i,j}_{k+1} w^{s,j}_{k} log \left( \frac{w_k^{m,s}}{\mathcal{LB}\left[{\eta}\right]} \right),	
	\end{multline}
	\begin{multline} \label{eq:entropy upper bound}
		\mathcal{UB} \left[ c \left( b_{k+1}\right)\right] \triangleq \mathcal{UB} \left[\mathcal{H} \right] =   \frac{\eta^s w_k^{m,s}}{\mathcal{LB}\left[{\eta}\right]} \left[ \mathcal{H}^s - log (\eta^s) \right] \\
		- \frac{w_k^{m,s}}{\mathcal{LB}\left[{\eta}\right]} \sum_{i}^{|L|} \sum_{j}^{M_{k}^s}  \tilde{\zeta}^{i,j}_{k+1} w^{s,j}_{k} log \left( \frac{w_k^{m,s}}{\mathcal{UB}\left[{\eta}\right]} \right)
		- \gamma log \left( \frac{\gamma}{\left|L\right| |\neg M_k^s|} \right),
	\end{multline} \normalsize
\end{theorem}
\textit{where $\gamma \triangleq 1 - \frac{\eta^s w_k^{m,s}}{\mathcal{UB}\left[ \eta \right]}$.} 

Furthermore,
considering different levels of simplifications, i.e. adding belief components to $M_k^s$, these bounds become tighter. 
\begin{corollary} \label{cost converge corollary}
	\textit{Given a simplified belief $b_k^s$, the bounds developed in Theorem \ref{cost bounds theorem} converge to $\mathcal{H}$ when $M_k^s = M_k$}
	\begin{equation}
		\underset{M_k^s \rightarrow M_k}{\text{lim}} \mathcal{LB} \left[\mathcal{H} \right] = \mathcal{H} = \mathcal{UB} \left[\mathcal{H} \right].
	\end{equation}
\end{corollary}
Using basic log properties, it is not hard to show that these bounds can  be incrementally adapted if one chooses to add additional components to $M_k^s$ (see full
derivation in supplementary material \cite{Shienman22icra_Supplementary}).

\subsection{Bounding $\eta$} \label{eta bounds section}
In this section we derive the bounds $\mathcal{LB} \left[ \eta \right]$ and $\mathcal{UB} \left[ \eta \right]$ over $\eta$. 
We start by expressing $\eta$ using $\eta^s$.
\begin{theorem} \label{analytic eta theorem}
\textit{Given a simplified belief $b_k^s$ at time $k$, for every action $u_k$ and considered future observation $z_{k+1}$, the normalization term $\eta$ in \eqref{eq:myopic objective} 
	can be expressed by}
\begin{equation} \label{eq:analytic eta theorem}
	\eta = w_k^{m,s} \eta^{s}
	+ \sum_{i}^{|L|} \sum_{j}^{\neg M_k^s} \tilde{\zeta}_{k+1}^{i,j}  w^{j}_{k} . 
\end{equation}
\end{theorem}
We can now use Theorem \ref{analytic eta theorem} to derive bounds for $\eta$. 
\begin{theorem} \label{eta bounds theorem}
\textit{Given a simplified belief $b_k^s$ at time $k$, the measurement likelihood term $\eta$ in \eqref{eq:myopic objective} 
	is bounded by}
\begin{align} \label{eq:eta bounds}
	\mathcal{LB} \left[ \eta \right] &= \eta^s w_k^{m,s}, 
	\\
	\mathcal{UB} \left[ \eta \right] &= \eta^s w_k^{m,s} + \left(1 - w_k^{m,s} \right) \sigma \sum_{i}^{\left| L \right| } \alpha^i, \label{eq:eta_upper_bound}
\end{align}
\textit{where $\sigma \triangleq \text{max} \left( \mathbb{P}\left( Z_{k+1}|\beta_{k+1}^i, x_{k+1} \right) \right)$ and $\alpha^i \triangleq \mathcal{UB} \left[ \mathbb{P} \left( \beta_{k+1}^i | x_{k+1} \right) \right]$ is an indicator function.}
\end{theorem}

As in Theorem \ref{cost bounds theorem}, since we only consider a subset of hypotheses these bounds are also computationally more efficient to calculate and become tighter when adding belief components to $M_k^s$.

\begin{corollary} \label{eta converge corollary}
	\textit{Given a simplified belief $b_k^s$, the bounds developed in Theorem \ref{eta bounds theorem} converge to $\eta$ when $M_k^s = M_k$}
	\begin{equation}
		\underset{M_k^s \rightarrow M_k}{\text{lim}} \mathcal{LB} \left[\eta \right] = \eta = \mathcal{UB} \left[\eta \right].
	\end{equation}
\end{corollary}
Furthermore, these bounds can also be incrementally adapted if one chooses to add additional components to $M_k^s$. (see full
derivation in supplementary material \cite{Shienman22icra_Supplementary}). 

\subsection{Simulating future observations $Z_{k+1}$}\label{subsec:FutureObs}
Evaluating the objective function \eqref{eq:myopic objective}, as explained in \cite{Pathak18ijrr}, is usually performed in two steps: we first simulate future observations $Z_{k+1}$ by sampling from the measurement likelihood $\eta$ using the generative model \eqref{eq:motion and observation models}, and then calculate the measurement likelihood $\eta$ for each such observation.  

While previous works, either with a Maximum Likelihood (ML) assumption, e.g. \cite{Elimelech21ijrr}, \cite{ Kitanov19arxiv}, \cite{Shienman21ral}, or without ML assumption, e.g. \cite{Sztyglic21arxiv, Zhitnikov21arxiv}, all consider the likelihood terms $\eta$ and $\eta^s$ to be equal, to the best of our knowledge, we are the first to consider impact of simplification on the normalization term in the myopic case. 

Recall that in our proposed approach we only evaluate the bounds over $\eta$ for each future observation. However, for the bounds in \eqref{eq: obj explicit bounds} to hold, we have to make sure we integrate over the same set of observations as in \eqref{eq:myopic objective}.
 
To handle this issue, we propose propagating and sampling from the original belief rather than from the simplified belief. We note that, as explained in \cite{Pathak18ijrr}, the concept of simulating future observations is computationally not the same as calculating the measurement likelihood which requires marginalizing over all possible data associations realizations and states. As such, using the original belief to simulate future observations does not affect the computational complexity of our proposed approach. In future research we plan to investigate if a similar approach is also applicable in the non-myopic setting.

\section{RESULTS}
We evaluate the performance of our approach in a highly ambiguous environment comprising perceptually identical landmarks in different locations. Our prototype implementation uses the
GTSAM library \cite{Dellaert12tr}  with a python wrapper; all experiments were run on an Intel  
i7-7850 CPU running at 2200 GHz with 32GB RAM.
\vspace{-6pt}
\setlength{\belowcaptionskip}{-12pt}
\begin{figure} [!h]
	\hspace*{-4pt}
	\includegraphics[scale=0.29]{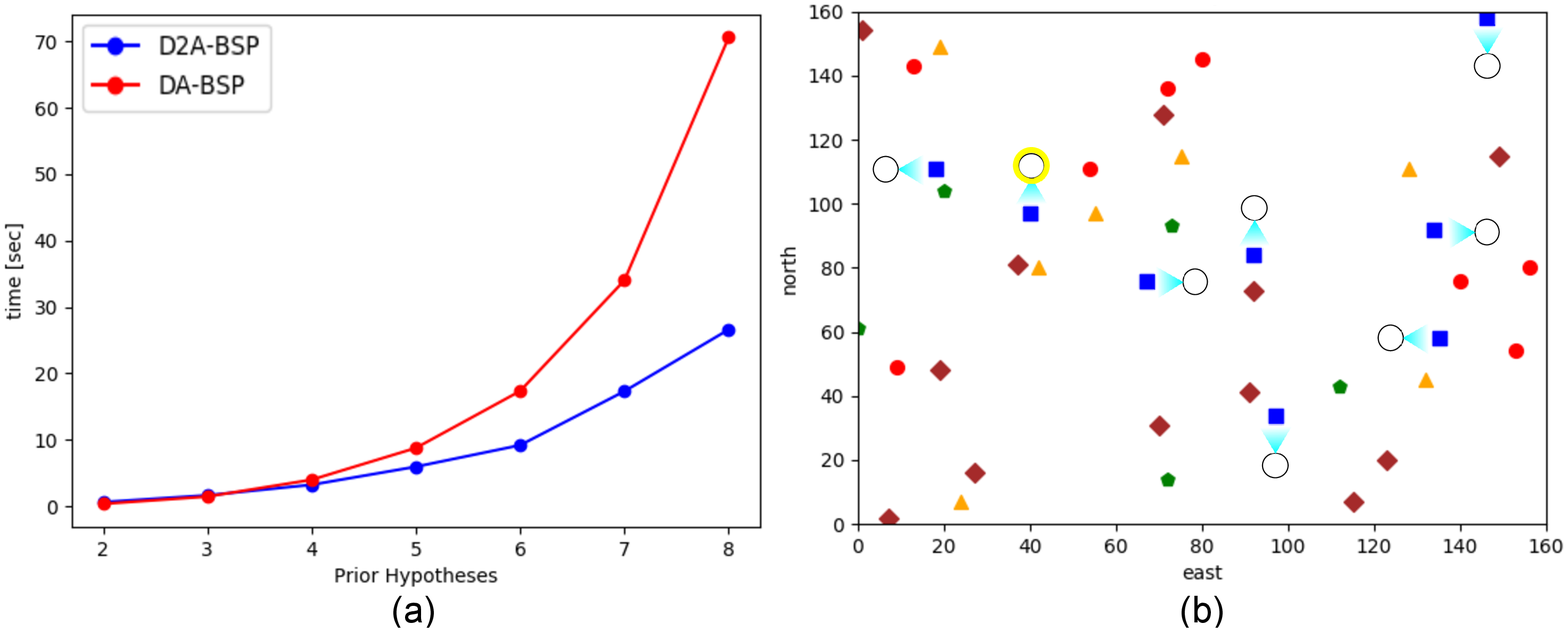}
	\scriptsize
	\caption{\scriptsize Given a multi-modal initial belief, the agent's goal is to fully disambiguate between all hypotheses. (a) Run-time [sec] as a function of number of prior hypotheses $M_0$; (b) A scenario with 8 prior hypotheses, each initialized in front of a blue square and denoted by a black ellipse. Headings are denoted with cyan triangles. The component that corresponds to the correct data association hypothesis, unknown to the agent, is highlighted in yellow.}
	\normalsize
	\label{fig: runtime experiment}
\end{figure}
\setlength{\belowcaptionskip}{0pt}

In our experiment we specifically consider five different landmark types represented by Squares, Circles, Diamonds, Pentagons and Triangles, randomly placed within the environment. The agent is initially placed in front of a blue square. With no other prior information, the initial belief is multi-modal containing $M_0$ hypotheses, each associated with a blue square. This scenario can be considered as a version of the kidnapped robot problem. 

The agent's goal is to fully disambiguate between hypotheses by solving the corresponding BSP problem (minimizing \eqref{eq:myopic objective}) at each planning session, considering entropy over posterior belief components weights as a cost function. The considered actions set at each planning session contains predefined motion primitives in all four cardinal directions. 

As we consider data association in inference as well, the number of belief components grows exponentially in time. For a fair comparison, we utilize the same pruning heuristics, based on a user defined weight threshold, for all approaches.

In Fig. \ref{fig: runtime experiment} we see the computational merits of \DDABSP{D2A-BSP} when there are no budget constraints. The higher the level of ambiguity within the environment, i.e. more hypotheses to reason about, the more prominent D2A-BSP becomes.
In this scenario, the distilled subset $M_k^s$ in each planning sessions is adapted greedily and incrementally, based on prior components weights, until \DDABSP{D2A-BSP} can guarantee the same action selection as DA-BSP.

Fig. 2 presents a scenario in which under hard budget constraints of $Q=6$, DA-BSP is unable identify the best action while \DDABSP{D2A-BSP} can. Recall that we only control the size of the prior belief, i.e. as $|L|=6$ only  one component can be used each time. As can be seen, using component $1$
DA-BSP$^Q$ (under budget) selects the action RIGHT which is clearly not the best action. The bounds of \DDABSP{D2A-BSP$^Q$} in this case are uninformative. However, using component $2$,
\DDABSP{D2A-BSP$^Q$} guarantees that LEFT is the best action (as bounds do not overlap). While DA-BSP$^Q$ yields LEFT as well in this case, it can only rely on heuristics to decide whether it should use component $1$ or component $2$.


\vspace{-10pt}
\setlength{\belowcaptionskip}{-10pt}
\begin{figure} [!h]
	\hspace*{-4pt}\includegraphics[scale=0.30]{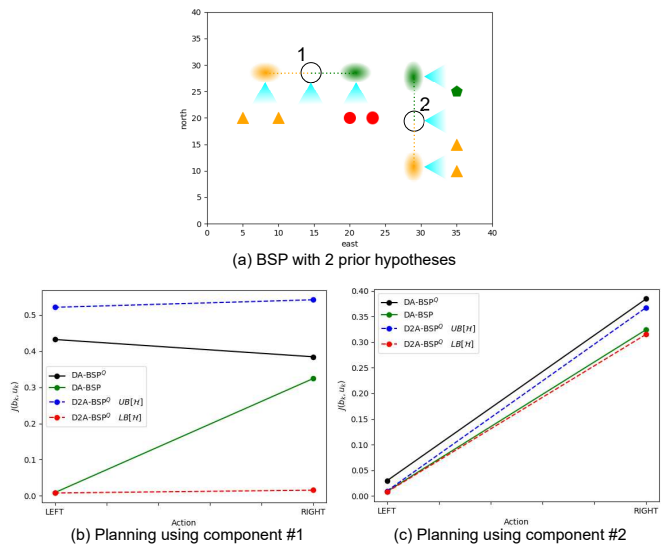}
	\caption{\scriptsize (a) A scenario under hard budget constraints of $Q=6$. The agent can only move LEFT or RIGHT.  Black ellipses and cyan triangles denote prior hypotheses and headings, respectively. Each prior weight equals $\frac{1}{2}$. Green and Orange ellipses denote the propagated belief after moving left or right, respectively. Moving LEFT is the best action as it is the only action that might lead to full disambiguation; (b) Objective function evaluations considering only component $1$ (denoted by superscript $Q$). DA-BSP in green represents the solution with no budget constraints considering all hypotheses; (c) Objective function evaluations considering only component $2$. Notations remain the same.}
	
	\label{fig: motivation figure}
\end{figure}
\setlength{\belowcaptionskip}{0pt}

\section{CONCLUSIONS}
In this work, we introduced a novel approach that utilizes a distilled subset of hypotheses to reduce the computational complexity in data association aware BSP with performance guarantees for the myopic case. While existing approaches handle the exponential growth of the number of hypotheses within planning using different heuristics which cannot provide performance guarantees,  we rigorously developed analytical bounds on the loss in quality of our proposed method solution. We then showed how to use these bounds in planning to obtain performance guarantees under hard budget constraints. We demonstrated our approach in an extremely aliased scenario where we were able to significantly reduce the computational complexity compared to existing approaches. 

Future work will consider a non-myopic case. Simplifying with respect to the number of considered landmarks in planning is another lucrative research direction as it directly affects the exponential growth in the number of hypotheses with the planning horizon.

\section{APPENDIX}
\subsection{Proof of Corollary \ref{corollary 1}}
We follow a similar derivation to the one presented in \cite{Pathak18ijrr} and factorize $w^{i,j}_{k+1}$ by first marginalizing over $x_{k+1}$ and then by applying the Bayes rule
\scriptsize
\begin{multline} 
	\!\!\!\!\!\!\! w^{i,j}_{k+1} \! = \!\!\!\! \int\displaylimits_{x_{k+1}} \!\!\!\! \frac{\mathbb{P}(Z_{k+1}|\beta_{k+1}^i,\beta_{1:k}^j, x_{k+1},H_{k+1}^{-}) \mathbb{P}(\beta_{k+1}^i, \beta_{1:k}^j, x_{k+1}|H_{k+1}^{-})}{\mathbb{P}(Z_{k+1}|H_{k+1}^{-})}.\nonumber
\end{multline} \normalsize
Using the chain rule multiple times over the second term in the numerator completes the proof.
$\hfill \blacksquare$

\subsection{Proof of Theorem \ref{analytic cost theorem}}
We split \eqref{entropy full} based on belief components from $M_k^s$ and use \eqref{eq: simplified belief mixture} to rewrite $\mathcal{H}$ as
\small
\begin{multline} \label{eq:entropy2}
	\mathcal{H} = - \sum_{i}^{|L|} \sum_{j}^{M_{k}^s} \frac{\tilde{\zeta}^{i,j}_{k+1}  w^{s,j}_{k}  w_k^{m,s}  }{\eta} log \left(\frac{\tilde{\zeta}^{i,j}_{k+1}  w^{s,j}_{k}  w_k^{m,s}}{\eta} \right) \\ - \sum_{i}^{|L|} \sum_{j}^{\neg M^{s}_{k}} \frac{\tilde{\zeta}^{i,j}_{k+1}  w^{j}_{k}}{\eta}  log \left(\frac{\tilde{\zeta}^{i,j}_{k+1} w^{j}_{k}}{\eta} \right).
\end{multline}
\normalsize
Using the key observation that $\tilde{\zeta}_{k+1}^{s,ij} = \tilde{\zeta}_{k+1}^{i,j}$, basic log properties and that by definition all posterior weights sum to 1, we write the cost for a simplified belief as
\small
\begin{equation} \label{eq:entropy simplified 2}
	\mathcal{H}^s = - \frac{1}{\eta^s} \sum_{i}^{|L|} \sum_{j}^{M^{s}_{k}} \left[ \tilde{\zeta}^{i,j}_{k+1}  w^{s,j}_{k} log \left(\tilde{\zeta}^{i,j}_{k+1} w^{s,j}_{k} \right) \right] + log \left( \eta^s \right).
\end{equation}
\normalsize
Replacing \eqref{eq:entropy simplified 2} back into \eqref{eq:entropy2} and using basic log properties completes the proof. $\hfill \blacksquare$

\subsection{Proof of Theorem \ref{cost bounds theorem}}
The last term in \eqref{eq:theorem 3} is non negative as all posterior weights are at most 1 by definition. Thus, removing this term and using theorem \ref{eta bounds theorem} we immediately get the lower bound. For the upper bound, we revisit the last term in \eqref{eq:theorem 3}. We first define $\gamma$ using \eqref{eq: simplified belief mixture} and \eqref{eq:eta s with proper zeta}
\begin{equation*}
	\gamma \triangleq \sum_{i}^{|L|} \sum_{j}^{\neg M^{s}_{k}} \frac{\tilde{\zeta}^{i,j}_{k+1}  w^{j}_{k}}{\eta} = 1 - \sum_{i}^{|L|} \sum_{j}^{M^{s}_{k}} \frac{\tilde{\zeta}^{i,j}_{k+1}  w^{j}_{k}}{\eta} = 1 - \frac{\eta^s   w^{m,s}_{k}}{\eta}.
\end{equation*}
Using the log sum inequality \cite{Cover91book}
\begin{equation*} \label{log sum inequality}
	\sum_{i}^{n} a_i \cdot log \left( \frac{a_i}{b_i} \right) \geq a \cdot log \left( \frac{a}{b} \right) \text{where} \sum_{i}^{n} a_i = a, \sum_{i}^{n} b_i = b,
\end{equation*}
with $a_i = \frac{\tilde{\zeta}_{k+1}^{i,j} w_k^j}{\eta}$ , $\sum\limits_{i}^{|L|} \sum\limits_{j}^{\neg M^{s}_{k}} \frac{\tilde{\zeta}^{i,j}_{k+1}  w^{j}_{k}}{\eta} = \gamma$ and $b_i=1$, we bound the last term in \eqref{eq:theorem 3}
\small
\begin{equation} \label{eq:not entropy upper}
	\sum_{i}^{|L|} \sum_{j}^{\neg M_k^s} \frac{\tilde{\zeta}_{k+1}^{i,j} w_k^j}{\eta}  log \left( \frac{\tilde{\zeta}_{k+1}^{i,j} w_k^j}{\eta} \right) \geq \gamma log \left( \frac{\gamma}{\left|L\right| |\neg M_k^s|} \right).
\end{equation} \normalsize
Substituting \eqref{eq:not entropy upper}  into \eqref{eq:theorem 3} and using Theorem \ref{eta bounds theorem} completes the proof.
$\hfill \blacksquare$

\subsection{Proof of Corollary \ref{cost converge corollary}}
Given $M_k^s=M_k$ it holds by definition that $w_k^{m,s}=1$ and $\mathcal{H} = \mathcal{H}^s$ as $b_{k+1}=b_{k+1}^s$. Substituting these back into 
\eqref{eq:entropy lower bound} and using  \eqref{eq: simplified belief mixture} and Corollary \ref{eta converge corollary}, the lower bound becomes
\small
\begin{multline}
	\mathcal{LB} \left[ \mathcal{H} \right] =   \mathcal{H} - log (\eta)
	-  \sum_{i}^{|L|} \sum_{j}^{M_k} \frac{\tilde{\zeta}^{i,j}_{k+1} w^{s,j}_{k}}{\eta}  log \left( \frac{1}{\eta} \right) = \mathcal{H}.
\end{multline}
\normalsize
Given $M_k^s=M_k$ it is also straightforward by Corollary \ref{eta converge corollary} that $\gamma = 0$. As such, using exactly the same derivations as for the lower bound, it immediately holds that $\mathcal{H} = \mathcal{UB}\left[ \mathcal{H} \right]$. This completes the proof. $\hfill \blacksquare$

\subsection{Proof of Theorem \ref{analytic eta theorem}}
We write $\eta$ explicitly and first marginalize over all data association realizations and states at time $k+1$. We then marginalize over all hypotheses from time $k$ and apply the chain rule multiple times
\smaller
\begin{multline} \label{eq:eta explicit}
	\!\!\!\!\!\!\! \eta \triangleq \mathbb{P} \left( Z_{k+1}|H_{k+1}^{-} \right) = \sum_{i}^{|L|} \sum_{j}^{M_k} \int\displaylimits_{x_{k+1}} \mathbb{P} \left( Z_{k+1} | x_{k+1} \right) 
	\mathbb{P} \left( \beta^i_{k+1} | x_{k+1} \right) \cdot \\   \mathbb{P} \left( x_{k+1} | \beta_{1:k}^j, H_{k+1}^{-} \right) \mathbb{P} \left( \beta_{1:k}^j | H_{k+1}^{-} \right) 
	=\sum_{i}^{|L|} \sum_{j}^{M_k} \tilde{\zeta}_{k+1}^{i,j}  w_k^j.
\end{multline} \normalsize
Using similar derivations and  the key observation that $\tilde{\zeta}_{k+1}^{s,ij} = \tilde{\zeta}_{k+1}^{i,j}$, we also write $\eta^{s}$ as 
\begin{equation} \label{eq:eta s with proper zeta}
	\eta^{s} = \sum_{i}^{|L|} \sum_{j}^{M_k^s} \tilde{\zeta}_{k+1}^{i,j}  w_k^{s,j}.
\end{equation}
Splitting \eqref{eq:eta explicit} based on belief components from $M_k^s$ and using \eqref{eq: simplified belief mixture} and \eqref{eq:eta s with proper zeta} completes the proof.
$\hfill \blacksquare$

\subsection{Proof of Theorem \ref{eta bounds theorem}}
As all weights are positive by definition, removing the last term in  \eqref{eq:analytic eta theorem} we immediately get the lower bound. For the upper bound, we 
 rewrite the second term in \eqref{eq:analytic eta theorem} using $\tilde{\zeta}_{k+1}^{i,j}$
\small
\begin{multline} \label{eq:explicit unnormalized sum}
	\sum_{i}^{|L|} \sum_{j}^{\neg M_k^s} \tilde{\zeta}_{k+1}^{i,j}  w^{j}_{k}
	= \sum_{j}^{\neg M_k^s} w^{j}_{k} \sum_{i}^{|L|} \int\displaylimits_{x_{k+1}} \mathbb{P}\left( Z_{k+1}|\beta_{k+1}^i, x_{k+1} \right) \cdot \\  \mathbb{P} \left( \beta_{k+1}^i|x_{k+1} \right)  
	\mathbb{P} \left(x_{k+1}|H_{k+1}^{-},\beta_{1:k}^j \right).
\end{multline}
\normalsize
The joint measurement likelihood term, given in \eqref{eq:joint measurement likelihood}, is a product of probability distribution functions, all given by \eqref{eq:motion and observation models} and can thus be bounded using an a priori known maximum value $\sigma$. The term $\mathbb{P} \left( \beta_{k+1}^i | x_{k+1} \right)$ represents the probability for the $i$th data association realization given $x_{k+1}$, i.e. the probability of observing a specific set of landmarks. As we assume the map to be given, it can be bounded using some constant $\alpha^i$, e.g. in the case of a camera, it can be an indicator function for landmarks that are within the field of view.
Finally, for every hypothesis $j$ it holds that $\int_{x_{k+1}}  \mathbb{P}  \left(x_{k+1}|H_{k+1}^{-},\beta_{1:k}^j \right) = 1$. Substituting these and $w_k^{m,s}$ back into \eqref{eq:analytic eta theorem}  completes the proof. $\hfill \blacksquare$

\subsection{Proof of Corollary \ref{eta converge corollary}}
Given $M_k^s=M_k$ it holds by definition that $w_k^{m,s}=1$ and $\eta = \eta^s$  as $b_{k+1}=b_{k+1}^s$. Replacing these back into \eqref{eq:eta bounds}, \eqref{eq:eta_upper_bound} immediately completes the proof.
 $\hfill \blacksquare$

\addtolength{\textheight}{-12cm}   


\bibliographystyle{plain}
\bibliography{refs}

\end{document}